\pdfoutput=1

\documentclass[11pt]{article}

\usepackage[]{authblk}
\usepackage{mdframed}
\usepackage{setspace}
\usepackage{multirow}
\usepackage[table,xcdraw]{xcolor}
\usepackage{float}
\usepackage[normalem]{ulem}
\usepackage{tipa}
\usepackage{array}
\usepackage{arydshln}

\makeatletter
\renewcommand\AB@affilsepx{, \protect\Affilfont}
\makeatother

\usepackage{tikz}
\usepackage[export]{adjustbox}

\usepackage[]{ACL2023}

\usepackage{times}
\usepackage{latexsym}
\usepackage{graphicx}
\usepackage{subcaption}

\usepackage{tabu}
\usepackage{multirow}
\usepackage{makecell} 

\usepackage[T1]{fontenc}

\usepackage[utf8]{inputenc}

\usepackage{microtype}

\usepackage{inconsolata}
\usepackage{csquotes}

\usepackage{tikz}

\usepackage{todonotes}
\usepackage{fdsymbol}

\newcommand{\pz}{\hphantom{0}}
\newcommand{\pzz}{\hphantom{00}}

\usepackage{fontawesome5}
\usepackage{overlays}

\title{Into the Unknown:\\ Generating Geospatial Descriptions for New Environments 
}
\author[1]{Tzuf Paz-Argaman}
\author[2]{John Palowitch}
\author[2]{Sayali Kulkarni}
\author[1,2]{\authorcr Reut Tsarfaty}
\author[2]{Jason Baldridge}
\affil[1]{Bar-Ilan University} 
\affil[2]{Google Research}
\affil[ ]{\authorcr \tt \{tzuf.paz-argaman, reut.tsarfaty\}@biu.ac.il}
\affil[ ]{\authorcr \tt \{palowitch, sayali, jasonbaldridge\}@google.com}

\date{}

\begin{document}

\maketitle

\begin{abstract}

Similar to vision-and-language navigation (VLN) tasks that focus on bridging the gap between vision and language for embodied navigation, the new \emph{Rendezvous} (RVS) task requires reasoning over allocentric spatial relationships (independent of the observer's viewpoint)  using non-sequential navigation instructions and maps. 
However, performance substantially drops in new environments with no training data.
Using opensource descriptions paired with coordinates (e.g., Wikipedia) provides training data but suffers from limited spatially-oriented text resulting in low geolocation resolution.
We propose a large-scale augmentation method for generating high-quality synthetic data for new environments using readily available geospatial data. Our method constructs a \emph{grounded knowledge-graph},
 capturing entity relationships. Sampled entities and relations (``shop north of school'') generate navigation instructions via
(i) generating numerous templates using context-free grammar (CFG) to embed specific entities and relations; (ii) feeding the entities and relation into a large language model (LLM) for instruction generation.
A comprehensive evaluation on RVS, showed that our approach improves the 100-meter accuracy by 45.83\% on unseen environments.
Furthermore, we demonstrate that models trained with CFG-based augmentation achieve superior performance compared with those trained with LLM-based augmentation, both in unseen and seen environments. These findings suggest that the potential advantages of explicitly structuring spatial information for text-based geospatial reasoning in previously unknown, can unlock data-scarce scenarios.

\end{abstract}

 \begin{figure}[!t]
  \centering
  \begin{minipage}[b]{0.48\textwidth}

            \includegraphics[width=1 \textwidth]{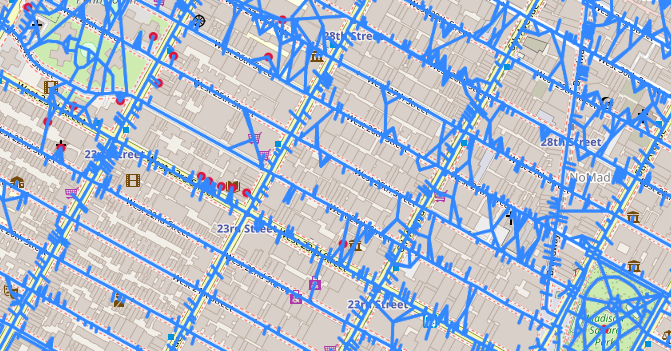}
              \end{minipage}
               \hfill
  \begin{minipage}[b]{0.48\textwidth}
    \includegraphics[width=\textwidth]{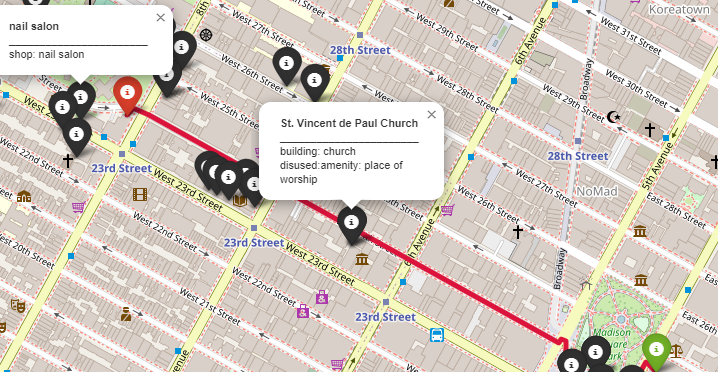}
  \end{minipage}

              \vspace*{-1mm}
                 {\footnotesize{ 
                   
                 \begin{spacing}{0.5}
                
                 \begin{flushleft} 
 \begin{mdframed}
 
\textit{\textbf{Meet me in the garden. Head northwest from St. Vincent de Paul Church for 2 intersections. The garden is right next to a fast-food restaurant. If you reach a nail salon, you have gone too far.
 } }
 \end{mdframed}
 \end{flushleft} 
\end{spacing} }}

        \caption
        { 
        Our method for generating spatial descriptions samples from the graph-map (top) a path (middle image, red line), a starting point (green marker), a goal point (red marker), and prominent landmarks (black markers). It then generates an instruction (bottom) from the spatial relations between these entities.}

        \label{fig:main_example}
    \end{figure}


\section{Introduction}

The ability to extract locations and paths from natural language descriptions of spatial information holds immense significance. This capability proves crucial in daily and disaster response scenarios, aiding the billions globally lacking formal addresses \cite{upu:2012, abebrese2019implementing, hu2023geo}, and enhancing Geographic Information Retrieval (GIR), particularly leveraging web-based resources \citep{spink2002sex, sanderson2004analyzing}.

Echoing the vision-and-language navigation (VLN) task's goal, of bridging the gap between visual perception and natural language (NL) instructions for embodied agents \cite{ku2020room}, the recently introduced Rendezvous (RVS) navigation task \cite{RVS} seeks to achieve a similar connection, but specifically between map representations and natural language NL. While VLN focuses on navigating within an environment based on visual cues and sequential instructions, RVS emphasizes the ability to utilize map information and non-sequential, often allocentric language descriptions to reach a specific target location. This shift from vision-centric instructions to map-aided guidance presents unique challenges, including reasoning about multiple spatial relationships simultaneously, inferring implicit actions from language, and navigating without explicit verification or step-by-step instructions.



However, there is a substantial gap between current models and the human performance on the RVS task, particularly in new environments that lack human annotated data.  
One approach to address this issue is to leverage naturally-occurring open-source data such as Wikipedia. However, these sources lack direct spatial information, which can result in models’ low performance ~\cite{solaz2023transformer}. Synthesizing data using large language models (LLMs) is a common method for addressing data scarcity in NLP  \citep{yoo2021gpt3mix, edwards2021guiding}.
However, for multimodal scenarios requiring precise spatial relationships, accurately generating such data without introducing errors or \enquote{hallucinations} (i.e., spurious relationships or entities), which severely undermines the performance on the underlying downstream task, remains a significant challenge.


We propose a method for generating high-quality  synthetic data for new environments using open-source geospatial data (Figure \ref{fig:main_example}). Our method constructs a grounded knowledge-graph of the environment, capturing spatial relationships between entities. By sampling these entities, abstract shapes like `blocks' (implicitly derived from street relationships), and relations (e.g., `the garden is next to a restaurant'), we generate navigation instructions by either  (i) creating a large amount
of templates via a generative context-free grammar (CFG), in which we embed the precise entities and  relations sampled, or (ii) feeding the entities and relation into an LLM which generates the instructions.

Extensive evaluation on the RVS  dataset demonstrates the clear advantage of our CFG-based method compared to the LLM approach. When navigating unseen environments, our method achieves a remarkable 9.1\% absolute increase in 100-meter accuracy and a substantial 39-meter decrease in median distance error. Overall, our method helps close the human-AI performance gap in unseen environments by  45.83\% in 100m accuracy and a decrease of 1,183m in median distance error. For the seen environment, our method results in an absolute improvement of 19.56\% in 100m accuracy and a decrease of 151m in median distance error.

\begin{figure*}[th]
\centering
\scalebox{0.99}{
\includegraphics[width=\textwidth]{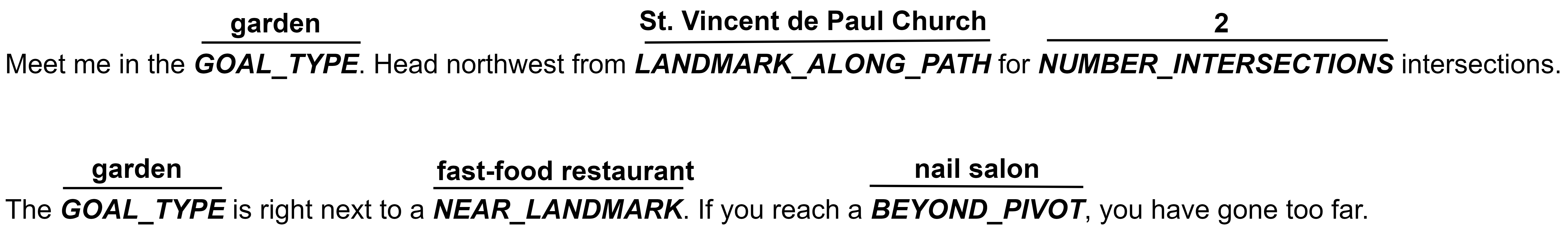}}

 \caption{Instruction generation steps (for example presented in Figure \ref{fig:main_example}): (i) Template creation via CFG; (ii) 
 replacing generic elements (in capital letters) with specific landmarks and spatial relations (above the lines).
 } 
 \label{fig:CFG}
\end{figure*}

\section{The Task}
The Rendezvous (RVS, \citealp{RVS}) task evaluates a system's ability to follow human-generated, colloquial language navigation instructions within a dense urban environment depicted by a map. The system is provided with three inputs:
(i) Detailed Map as Knowledge Graph: a comprehensive map of the environment represented as a knowledge graph. This graph encodes spatial relationships between landmarks and other relevant features.
(ii) Explicit Starting Point (Geo-coordinates): the starting location specified as a latitude and longitude coordinate pair.
(iii) Navigation Instruction: a natural language instruction describing the target location's relative position to landmarks and the starting point. This instruction leverages colloquial language typically used in navigation scenarios.
The RVS task demands the system to process these inputs and generate the goal location's coordinates within the defined map boundaries.

\section{Proposed: Relational Augmentation}
\label{sec:augmentation}

Our augmentation method aims to generate natural language location descriptions that are both accurate and well grounded. To that end, we leverage  \href{http://www.openstreetmap.org}{OpenStreetMap (OSM)}.\footnote{OpenStreetMap is a user-updated map of the world -- http://www.openstreetmap.org} It involves three stages: (i) sampling paths; (ii) calculating spatial relations between entities; and (iii) generating instructions based on the spatial relations calculated in stage (ii).
We use the OSM-based graph provided in RVS for the first and second parts and provide two methods for generating instructions based on the spatial relations between entities calculated in that part.

\subsection{Sampling paths}
\label{Sec_sample_paths}


To generate RVS-like samples accurately, we follow the RVS sampling protocol. We randomly sample small entities for the end point (the entity’s shape has a maximum radius of 100 meters). For the start point, we randomly select an entity that (i) is within 200–2000 meters of the end point; (ii) has a name or type tag (e.g., a bookshop). This information allows us to refer to the entity in the instruction by its definite description and not by its proper name.
Finally, we pick a route that is the shortest distance between the start and end points.

\subsection{Calculating Spatial Relations between Entities}
\label{sec_calculating_spatial}

Leveraging the Open Street Map (OSM) graph, we identify prominent landmarks relevant to the sampled paths for inclusion in navigation instructions. 
 However, not all paths necessarily contain all types of landmarks and features. For instance, if the navigation ends in a dead-end street, landmarks beyond the leading path might not be relevant to the task. This section details the criteria for landmark selection and the spatial relations computed with respect to the start and end points, and the path.


\paragraph{Picking landmarks}
We pick three different types of landmarks with a certain spatial relation that will be referenced in the instruction: (i) landmarks close to the end point (within 100 meters from it); (ii) landmarks along the route; and (iii) landmarks that are on the same street as the goal location but beyond the route, such that if the agent keeps on walking beyond the goal, it will reach that landmark (“beyond landmark”). Priority for landmark selection is given to those with the highest level of external recognition, as determined by the following hierarchy: has a Wikipedia 
or Wikidata\footnote{Wikidata is Wikipedia's free, open, and interconnected knowledge base. \href{https://www.wikidata.org/}{https://www.wikidata.org/}} link, is a brand, is a tourism attraction, is an amenity, is a shop. We randomly select all landmarks from the most prominent level found. If in the area there are multiple landmarks of the same type, we group the landmarks according to their type and quantity (e.g., `two book shops'). If a landmark is far (over 200 meters) from the end point and it has a proper name, we can use its proper name in the text generation, e.g., ‘the Empire State Building’. If it is near (less than 200 meters) the end point, we will always use the indefinite name of the landmark, e.g., ‘a bookshop’.

\paragraph{Calculating spatial relations}

The objective is to determine the spatial relationship between landmarks and the end point. There are several ways to describe the relationship between two entities, such as the number of blocks between them. The spatial relations calculated are (i) allocentric relations, i.e., cardinal directions, between landmarks, start and end points. Cardinal direction is calculated by the bearing $\theta_{\text{degrees}}$ between two points (longitude, latitude), $(x_1,y_1)$ and $(x_2,y_2)$:

\begin{equation}
\resizebox{.92\linewidth}{!}{$
\begin{aligned}[fleqn]
\label{Eq_bearing}
\theta_{\text{radians}} &= (\tan^{-1} \left( \sin(\lambda_2 - \lambda_1) \cos(\varphi_2), \right. \\
&\qquad \left. \cos(\varphi_1) \sin(\varphi_2) \right) \\
&\qquad \left. - \sin(\varphi_1) \cos(\varphi_2) \cos(\lambda_2 - \lambda_1) \right) \\
&\qquad \left. + 360^\circ)\mod 360^\circ \right) \\
\theta_{\text{degrees}} &= \theta_{\text{radians}} \cdot \frac{180^\circ}{\pi}
\end{aligned} $}
\end{equation}

Where $\lambda_1 = \frac{x_1 \cdot\pi}{180^\circ}$,  
$\lambda_2 = \frac{x_2 \cdot\pi}{180^\circ}$, $\varphi_1 = \frac{y_1 \cdot\pi}{180^\circ}$ and $\varphi_2 = \frac{y_2 \cdot\pi}{180^\circ}$. 
Bearings $\theta_{\text{degrees}}$ fall into different ranges, each with a corresponding cardinal direction (e.g., `North-West').
(ii) Egocentric relations between landmarks and end point to the path, e.g., `on the right side'. 
To calculate egocentric relations, we rely on two key angles (Eq. \ref{Eq_bearing}): the bearing of the path itself $\theta^p_{\text{degrees}}$, and the bearing of the shortest imaginary line connecting the path to the landmark's point 
 $\theta^l_{\text{degrees}}$:
\vspace{-0.2cm}
\begin{equation}
\begin{aligned}[fleqn]
\label{Eq_egocentric}
\Delta \theta^{l-p} &= (\theta^l_{\text{degrees}} - \theta^p_{\text{degrees}})\mod 360^\circ \\
\end{aligned}
\end{equation}
\begin{flalign*}
\begin{cases}
\text{`RIGHT'}, & \text{if } \Delta \theta^{l-p} < 180^\circ \\
\text{`LEFT'}, & \text{otherwise}
\end{cases}
\end{flalign*}
`LEFT' and `RIGHT' indicate the landmark’s position relative to the path.
(iii) The number of blocks and intersections the agent must pass through to reach the end point. (iv) the end point's egocentric and allocentric position on the block, e.g., `middle of the block' and `north-east corner of the block'. 
Allocentric position on the block involves determining the bearing (Eq. \ref{Eq_bearing}) of the path along the block and mapping it to cardinal direction as in (i).

\subsection{Data Generation}

Based on the sampling and spatial relations' calculations we use two methods to generate the instruction: via templates created with a CFG
~\cite{chomsky1956three}, and via prompting an LLM --- for enlarging the vocabulary and the style of the text.


\paragraph{CFG-based Method}

Here the key idea is using a CFG to  generate templates that can then be adapted according to the sampled data. The CFG we design requires defining terminal symbols (lexical elements), nonterminal symbols, and production rules. The nonterminals contain the main parts of what a path description contains, such as descriptions around the goal, along the path, what to avoid, and so on. The terminals contain optional variations, for example, the verb for the agent to proceed can be `go', `walk', and so on. The grammar creates templates that are processed into instructions, as demonstrated in Figure \ref{fig:CFG}. For a given sampled path, we randomly pick a template that contains all the landmark categories and spatial relations calculated for the path.
We then generate the instruction by replacing the variables in the chosen template with the corresponding landmarks and spatial relations (e.g., `\textit{NUMBER\_INTERSECTIONS}' will be replaced with the actual number of intersections the agent should walk).

\paragraph{Prompting LLMs}
Using the aforementioned template-based instructions, we prompted an LLM to `rephrase the subsequent navigation instruction, ensuring it explains how to travel from the starting position to the destination: \textit{Navigation Instruction}', where the \textit{Navigation Instruction} is an instruction based on the CFG generation process. For example, based on the example in Figures \ref{fig:main_example} and \ref{fig:CFG}  we get the following sentence: `Head northwest from St. Vincent de Paul Church for 2 intersections. The garden is next to a fast-food restaurant.

\section{Experimental Setup}



\subsection{Evaluation}
We follow the RVS evaluation metrics:  (i) 100m accuracy; (ii) 250m accuracy for coarse-grained evaluation; (iii) mean absolute error distance (MAE); (iv) median absolute error (Med.AE); (v) maximum absolute error (Max.AE);
and (vi) area under the curve (AUC) of the error distance.
Here are the formulas for evaluating set S with metrics (iii-vi):
\begin{equation}
\resizebox{.89\linewidth}{!}{$
\text{MAE(S)} = \frac{1}{|S|} \sum_{s\in S} dist(loc(s), approx(s))
$}
\end{equation}
\vspace*{-\baselineskip}
\vspace*{-\baselineskip}

\begin{equation}
\resizebox{.89\linewidth}{!}{$
\text{Med.AE(S)} = \{dist(loc(s), approx(s))| s\in S\}_{ \lfloor |S|/2 \rfloor}
$}
\end{equation}

\vspace*{-\baselineskip}
\begin{equation}
\resizebox{.89\linewidth}{!}{$
\text{Max.AE(S)} = max(\{dist(loc(s), approx(s))| s\in S\}) 
$}
\end{equation}
\vspace*{-\baselineskip}
\vspace*{-\baselineskip}

\begin{equation}
\resizebox{.89\linewidth}{!}{$
\text{AUC(S)} = 
\frac{\int_0^\infty (\log{dist(loc(s), approx(s))+ \epsilon| s\in S)\uparrow }  ds}{\log{H_{max}} \cdot (|S| - 1)}
$}
\end{equation}

Where $\epsilon=1e-5$ and $H_{max}=20,037\cdot10^3$, approximately
the maximum haversine distance.

\subsection{Models for evaluation}
\paragraph{T5-model} 
We test our augmentation method with T5 model, a transformer-based encoder-decoder model designed with a text-to-text format ~\cite{raffel2020exploring}. Both encoder and decoder utilize multi-head, multi-layer self-attention mechanisms ~\cite{vaswani2017attention}. Given input sequence text $X = (x_1, ..., x_N)$ and a starting point $p_s$, the encoder encodes the instruction and the starting point's representation such that $E^l = (e^l_1,..., e^l_N, e^{l}_{p_s})$ where $l\in L$, representing the $L$ hierarchical encoded layers. The output of the final encoder layer is a sequence of hidden vectors $H = (h_1, ..., h_N,h_{p_s})$.
The decoder generates output tokens sequentially, predicting the probability $p(p_t | p_{1:t-1}, H) = \operatorname{softmax}(W_o \otimes h'_t)$ of token  $p_t$ at step $t$, based on the previous outputs and hidden state $h'_t$.
Importantly, the model is trained with a pre-defined high-level path $P$ that guides the generation process. This path starts at the starting point, traverses through prominent landmarks ordered by their direction relative to the goal, and eventually reaches the goal itself.

\paragraph{Non-learning {\scshape Landmark}  Baseline} Predicts the location of a prominent landmark (defined in Sec. \ref{sec_calculating_spatial}) in the map within a radius of 1 kilometer.

\begin{table}[t]
\centering
\scalebox{0.65}{
\begin{tabular}{lllll}
                 & RVS   & Aug-CFG & Aug-Prompt & Aug-WikiGeo \\ \hline
Avg. Text Length & 43.47 & 33.70    & 37.26      & 24.82       \\
Avg.Entities     & \pz3.98  & \pz4.01    & \pz4.00          & \pz2.20        
\end{tabular}
}
\caption{ Statistics over RVS, and augmentation data.
}
\label{tab:data_stat}
\end{table}

\begin{table*}[th]
\centering
\scalebox{0.6}{

\begin{tabular}{llccccccc}
& \textbf{Method}   & \textbf{Training Set}   &  \textbf{100m Accuracy} & \textbf{250m Accuracy}         & \textbf{MAE}                                        & \textbf{Med.AE} & \textbf{Max.AE}  & \textbf{AUC} \\ 

\cline{2-9} 
&
\multicolumn{8}{c}{\cellcolor[HTML]{EFEFEF}\textbf{Manhattan (Manh) Seen-city Development Results}}                                                                                                                         \\ \cline{2-9} 
{\tiny 1} & \textbf{{\scshape Human} } & NA     & 88.12                   & 95.64                            & \pzz\pz74                                                      & \pzz\pzz4                  & 2,996              & 0.10                  \\
{\tiny 2} &
\textbf{{\scshape Landmark}}     & NA & 0.54                   & 5.26                            & \pzz776                                                   & \pzz815              & 1,384            & 0.39                  \\

{\tiny 3} & \textbf{{\scshape T5} }  & RVS Train-set      & \pzz\pzz 27.92 \small{(0.39)}           
& \pzz\pzz52.63 \small{(0.45)}
& \pzz\pzz\pz 362 \small{(9)}                                             & \pzz\pzz 231 \small{(3)}       & \pzz\pzz2,957 \small{(641)} & \pzz\pzz0.32 \small{(0.00)}           \\

 \hdashline
{\tiny 4} & \textbf{{\scshape T5}} & Aug-WikiGeo Manh & \pzz\pzz\pz 0.00 \small{(0.00)}         
& \pzz\pzz \pz 1.54  \small{(0.00)}                  

& \pzz\pz 1,085 \small{(0)}                                             & \pzz \pz 1,124 \small{(0)}       & \pzz 1,929 \small{(0)}   & \pzz\pzz0.41 \small{(0.00)}           \\

{\tiny 5} & \textbf{{\scshape T5}} & Aug-CFG Manh & \pzz\pzz 28.83 \small{(0.63)}         
& \pzz\pzz 46.15  \small{(0.77)}                  

& \pzz\pzz\pz 668 \small{(17)}                                             & \pzz\pzz \pz 304 \small{(27)}       & \pzz\pzz \pz 4,637 \small{(2,207)}   & \pzz\pzz 0.34 \small{(0.00)} 
\\

{\tiny 6} & \textbf{{\scshape T5}} & Aug-Prompt Manh & \pzz\pzz21.32 \small{(0.20)}         
& \pzz\pzz37.01  \small{(0.14)}

& \pzz\pzz\pz 963 \small{(17)}                                             & \pzz\pzz \pz658 \small{(14)}       & \pzz\pzz\pz 6,731 \small{(1,003)}   & \pzz\pzz0.36 \small{(0.00)}           \\

{\tiny 7} & \textbf{{\scshape T5}} & Aug-CFG Manh \& RVS Train-set & \pzz\pzz \textbf{45.97 \small{(1.34)}    }     
& \pzz\pzz \textbf{64.01  \small{(0.89)}  }                

& \pzz\pzz\pz 377 \small{(32)}                                            & \pzz\pzz \pz \textbf{121 \small{(15)} }      & \pzz\pzz  5,317 \small{(831)}   & \pzz\pz 0.3 \small{(0.00)} 
\\

\cline{2-9} 
&
\multicolumn{8}{c}{\cellcolor[HTML]{EFEFEF}\textbf{Pittsburgh (Pitt) Unseen-Development Results}}                                                                                                                      \\ \cline{2-9} 
{\tiny 8} & \textbf{{\scshape Human}} & NA     & 86.94                   & 92.94                            & \pzz\pz99                                                         & \pzz \pz7                     & 2,951               & 0.13                  \\

{\tiny 9} &   \textbf{{\scshape Landmark}} &   NA & {1.47}                    &      {9.48}                           &   \pzz  {677}                                                       &        \pz  {691}             &              1,345       &        0.38               \\

{\tiny 10} & \textbf{T5}    & RVS Train-set    & \pzz\pzz\pz 0.49 \small{(1.47)}            &               \pzz\pzz \pz 2.34 \small{(1.44)}                 & \pzz\pzz1,171 \small{(24)}                                                 & \pzz \pz1,107 \small{(14)}            & \pzz\pzz4,701 \small{(101)}         & \pzz\pzz 0.41 \small{(0.00)}           \\

 \hdashline
 
{\tiny 11} & \textbf{{\scshape T5}} & Aug-WikiGeo Pitt & \pzz\pzz\pz  0.00 \small{(0.00)}         
& \pzz\pzz \pz 2.05  \small{(0.00)}                  

& \pzz\pzz\pz 961 \small{(0)}                                             & \pzz\pzz 955 \small{(0)}       & \pzz\pzz1,912 \small{(0.00)}   & \pzz\pzz0.40 \small{(0.00)}           \\

{\tiny 12} & \textbf{{\scshape T5}} & Aug-CFG Pitt & \pzz\pzz\pz \textbf{46.63} \small{(0.54)}         
& \pzz\pzz \textbf{63.73}  \small{(0.41)}                  & \pzz\pzz\pz 466 \small{(5)}                                             & \pzz\pzz 120 \small{(1)}       & \pzz\pzz5,251 \small{(0.00)}   & \pzz\pzz0.31 \small{(0.00)}           \\

{\tiny 13} & \textbf{{\scshape T5}} & Aug-Prompt Pitt & \pzz\pzz\pz 37.10 \small{(0.49)}         
& \pzz\pzz58.30  \small{(0.34)}                  

& \pzz\pzz\pz 492  \small{(5)}                                             & \pzz\pzz  159 \small{(1)}       & \pzz 5,251 \small{(0)}   & \pzz\pzz0.32 \small{(0.00)}           \\

{\tiny 14} & \textbf{{\scshape T5}} & RVS Train-set \& Aug-CFG Pitt & \pzz\pzz\pz 46.24 \small{(0.30)}         
& \pzz\pzz 62.85  \small{(0.41)}                

& \pzz\pzz\pzz \textbf{387} \small{(17)}                                             & \pzz\pzz \textbf{116} \small{(4)}       & \pzz\pz 5,162 \small{(103)}   & \pzz\pzz0.30 \small{(0.00)}           \\

\cline{2-9} 

&
\multicolumn{8}{c}{\cellcolor[HTML]{EFEFEF}\textbf{Philadelphia (Phila) Unseen-city Zero-shot Results}}                                                                                                                             \\ \cline{2-9} 

{\tiny 15} & \textbf{{\scshape Human}} & NA    &  93.64                   & 97.97                            & \pzz\pz27                                                      & \pzz\pz3                  & 2,708            & 0.05                  \\

{\tiny 16} & \textbf{{\scshape Landmark}}  & NA    & \pz 1.02                   & \pz 7.90                            & \pzz707                                                   & \pz 713              & 1,384            & 0.38                  \\

{\tiny 17} & \textbf{{\scshape T5}} & RVS Train-set        & \pzz\pzz \pzz 0.26 \small{(0.05)}           
& \pzz\pzz \pz 1.80 \small{(0.27)}

& \begin{tabular}[c]{@{}c@{}}\pzz\pz1,362 \small{(43)}\end{tabular} & \pzz\pz1,308 \small{(35)}      & \pzz\pzz6,911 \small{(454)}  & \pzz\pzz0.42 \small{(0.00)}           \\

 \hdashline

{\tiny 18} & \textbf{{\scshape T5}} & RVS Train-set \& Aug-CFG Phila   & \pzz\pzz \textbf{46.09 \small{(0.50)}}         
& \pzz\pzz \textbf{61.66  \small{(0.00)} }                 

& \pzz\pzz \textbf{579 \small{(2)}}                                             & \pzz\pzz  \textbf{125 \small{(3)} }      & \pzz\pzz 5,774 \small{(715)}   & \pzz\pzz0.31 \small{(0.01)}          

\end{tabular}

}
\caption{Results are divided over RVS's test (Philadelphia) and development sets (Manhattan and Pittsburgh). 
The distance errors are presented in meters. For the learning models, we report the mean over three random initializations, and the standard deviation (STD) is in brackets. }
\label{tab:results}

\end{table*}

\subsection{Data}

\paragraph{RVS}
 The RVS \citep{RVS} dataset serves as a human-level benchmark for the purpose of evaluating the ability to follow allocentric navigation instructions based on a map. It consists of English navigation directives, each paired with a start and end point. The data is divided into four distinct sets:
(i) \textit{Training-set} -- containing 7,000 instructions from Manhattan; (ii) \textit{Seen-city development-set} -- containing 1,103 instructions from Manhattan; (iii) \textit{Unseen-city development-set} -- containing 1,023 instructions from Pittsburgh; (iv) \textit{Test-set} -- containing 1,278 instructions from Philadelphia.\\

\noindent
The following datasets are all synthetically generated. We generated 200,000 instructions per dataset for each region, with the exception of Aug-WikiGeo:
\paragraph{Aug-CFG} Data created with the CFG method described in Section \ref{sec:augmentation}. The CFG method created 194,721 templates with 15 production rules.
The vocabulary contains only unique 111 tokens. Table \ref{tab:data_stat} shows that this method produces shorter instructions than the RVS, but with more entities in each instruction. Figures \ref{fig:main_example} and \ref{fig:CFG} show an example of an instruction generated based on this method.

\paragraph{Aug-CFG-Allocentric} Data created with a CFG including templates with allocentric spatial relations between entities. E.g., `the school is north of the bar'. The data contains 69,720 templates.

\paragraph{Aug-CFG-Egocentric} Data created using a CFG which contains templates with egocentric spatial relations between entities, e.g., `the bar on your right'. The data contains 112,640 templates.

\paragraph{Aug-CFG-Minimal} This dataset leverages a minimal set of templates (64) to represent the full range of entities and spatial relationships found in the Aug-CFG data. We constructed Aug-CFG-Minimal using a greedy selection process. We began by selecting the first template. Subsequently, we added templates that introduced new spatial features not covered by the existing set. This process continued until we reached a final set of 64 templates that comprehensively capture all possible spatial features.


\paragraph{Aug-Prompt} Data generated by prompting  PaLM2 ~\cite{anil2023palm} with the prompting method described in Section \ref{sec:augmentation}.\footnote{We use PaLM2 `models/text-bison-001' - for more details, see \href{https://developers.generativeai.google/tutorials/text_quickstart}{https://developers.generativeai.google}}
Table \ref{tab:data_stat} shows that this method produces longer instructions than the CFG-based method but with fewer entities.



\begin{table*}[th]
\centering
\scalebox{0.62}{

\begin{tabular}{llccccccc}
& \textbf{Method}   & \textbf{Training Set}   &  \textbf{100m Accuracy} & \textbf{250m Accuracy}         & \textbf{MAE}                                        & \textbf{Med.AE} & \textbf{Max.AE}  & \textbf{AUC} \\ 

\cline{2-9} 
&
\multicolumn{8}{c}{\cellcolor[HTML]{EFEFEF}\textbf{Manhattan (Manh) Seen-city Development Results}}                                                                                                                         \\ \cline{2-9}

{\tiny 1} & \textbf{{\scshape T5}} & Aug-CFG Manh & \pzz\pzz 28.83 \small{(0.63)}         
& \pzz\pzz 46.15  \small{(0.77)}                  

& \pzz\pzz\pz 668 \small{(17)}                                             & \pzz\pzz \pz 304 \small{(27)}       & \pzz\pzz \pz 4,637 \small{(2,207)}   & \pzz\pzz 0.34 \small{(0.00)} 
\\

{\tiny 2} & \textbf{{\scshape T5}} & Aug-CFG-Allocentric Manh & \pzz\pzz 29.83 \small{(1.22)}         
& \pzz\pzz 47.96  \small{(2.95)}  

& \pzz\pzz\pz 681 \small{(39)}                                             & \pzz\pzz \pz 310 \small{(66)}       & \pzz\pz 6,446 \small{(88)}   & \pzz\pzz 0.34 \small{(0.01)} \\

{\tiny 3} & \textbf{{\scshape T5}} & Aug-CFG-Egocentric Manh & \pzz\pzz 28.83 \small{(0.38)}         
& \pzz\pzz 45.87  \small{(0.13)}  

& \pzz\pzz\pz 751 \small{(29)}                                             & \pzz\pzz  384 \small{(5)}       & \pzz\pz 4,572 \small{(77)}   & \pzz\pzz 0.35 \small{(0.00)} \\

{\tiny 4} & \textbf{{\scshape T5}} & Aug-CFG-Minimal Manh & \pzz\pzz 23.21 \small{(0.39)}         
& \pzz\pzz 39.62  \small{(0.13)}  

& \pzz\pzz\pz 880 \small{(12)}                                             & \pzz\pzz \pz 567 \small{(47)}       & \pzz\pzz\pz 5,463 \small{(2,228)}   & \pzz\pzz 0.36 \small{(0.00)} \\

{\tiny 5} & \textbf{{\scshape T5}} & Aug-Dummy Manh & \pzz\pzz 1.35 \small{(0.13)}         
& \pzz\pzz 5.62  \small{(0.13)}  

& \pzz\pzz\pz 1,240 \small{(28)}                                             & \pzz\pzz \pz 1,130 \small{(41)}       & \pzz\pzz\pz 4,654 \small{(0.00)}   & \pzz\pzz 0.41 \small{(0.00)} \\

\cline{2-9} 
&
\multicolumn{8}{c}{\cellcolor[HTML]{EFEFEF}\textbf{Pittsburgh (Pitt) Unseen-Development Results}}                                                                                                                      \\ \cline{2-9}

{\tiny 6} & \textbf{{\scshape T5}} & Aug-CFG Pitt & \pzz\pzz\pz \textbf{46.63} \small{(0.54)}         
& \pzz\pzz \textbf{63.73}  \small{(0.41)}                  

& \pzz\pzz\pz 466 \small{(5)}                                             & \pzz\pzz 120 \small{(1)}       & \pzz\pzz5,251 \small{(0.00)}   & \pzz\pzz0.31 \small{(0.00)}           \\

{\tiny 7} & \textbf{{\scshape T5}} & Aug-CFG-Allocentric Pitt & \pzz\pzz\pz 45.94 \small{(0.48)}        
& \pzz\pzz63.64  \small{(0.55)}                  

& \pzz\pzz\pzz 453 \small{(18)}                                             & \pzz\pzz 119 \small{(0)}       & \pzz\pz 4,835 \small{(91)}   & \pzz\pzz0.31 \small{(0.00)}           \\

{\tiny 8} & \textbf{{\scshape T5}} & Aug-CFG-Egocentric Pitt & \pzz\pzz\pz 43.01 \small{(0.48)}         
& \pzz\pzz60.22  \small{(1.60)}                  

& \pzz\pzz\pz 514 \small{(2)}                                             & \pzz\pzz 131 \small{(1)}       & \pzz\pzz6,034 \small{(399)}   & \pzz\pzz0.31 \small{(0.00)}           \\

{\tiny 9} & \textbf{{\scshape T5}} & Aug-CFG-Minimal Pitt & \pzz\pzz\pz 42.82 \small{(2.07)}         
& \pzz\pzz62.84  \small{(0.20)}                  

& \pzz\pzz\pzz 479 \small{(26)}                                             & \pzz\pzz 132 \small{(3)}       & \pzz\pzz\pz 3,160 \small{(1030)}   & \pzz\pzz0.31 \small{(0.00)}           \\

{\tiny 10} & \textbf{{\scshape T5}} & Aug-Dummy Pitt & \pzz\pzz \pzz 6.26 \small{(0.83)}         
& \pzz\pzz 13.88  \small{(1.04)}  

& \pzz\pzz\pz 1,069 \small{(6)}                                             & \pzz\pzz \pz 933 \small{(1)}       & \pzz\pzz\pz 3,197 \small{(746)}   & \pzz\pzz 0.39 \small{(0.00)} \\

\end{tabular}

}
\caption{CFG-based Augmentation Ablation Results 
 }
\label{tab:results_ablation_CFG}

\end{table*}

\paragraph{Aug-WikiGeo}
To facilitate a comparison with standard open-source location augmentation methods, such as Wikipedia-based  ~\cite{krause2023geographic}, we generated the comprehensive Wikigeo dataset by consolidating data from Wikipedia (pages and backlinks), Wikidata (pages), and OpenStreetMap (entities). The data for Manhattan (Manh., 255,663 samples), Pittsburgh (Pitt., 27,401 samples), and Philadelphia (Phila., 52,367 samples) are generated based on the same regions as in the evaluation. Table \ref{tab:data_stat} shows that this method produces shorter instructions and fewer entities than the CFG-based and prompting-based methods.

\paragraph{Aug-Dummy} To assess the contribution of the text augmentation, as opposed to just learning all possible paths from a known starting point, we use an augmentation where the text does not convey any spatial details of the location. To create the non-spatial text, we used PaLM2 by requesting it to rephrase the sentence \enquote{Meet me here} and got a total of 31 versions. Path sampling followed the method described in Section \ref{Sec_sample_paths}.

\section{Results}

\subsection{Analysis of Quantitative Results}


Table \ref{tab:results} shows the results of our experiments over RVS's seen-city development set (Manh.), unseen-city development set (Pitt), and the unseen-city test set (Phila.).
For the seen environment (Manh.), training on synthetic data only (line 5) outperforms human-annotated data (line 3). The gap is small, but if the model is first trained on synthetic data and then on human data (line 7), the gap is a 65\% ratio in 100m accuracy and a 110m lower in Med.AE.

Training on real human data from other regions (lines 10, 17) fails to translate to unseen environments like Pittsburgh and Philadelphia. 
However, injecting region-specific synthetic data (line 12) dramatically boosts performance: 46.14\% higher 100m accuracy and 987m lower Med.AE.
WikiGeo's emphasis on local features sacrifices spatial relational understanding, leading to lower performance on all development sets (lines 4, 11). However, its superior max distance estimation (up to 2km) and lower distance error over models trained on human-annotated data 
from a different region
(lines 11 vs. 10), suggest a strong ability to learn path boundaries based on localized information.


Aug-Prompt, while generating instructions with richer language and stylistic diversity compared to Aug-CFG's template-based approach, exhibits a marked decrease in performance (lines 6, 13 vs. lines 5, 12). Sampling 20 instructions from Aug-Prompt, we found that in two cases, the type of goal was emitted, and in five cases, the model `hallucinated' -- adding incorrect spatial relations. This indicates a potential trade-off between linguistic complexity and the fidelity of spatial information within generated instructions.

 \begin{figure*}[!ht]
        \centering
        \begin{subfigure}[b]{0.495\textwidth}
            \centering
            \includegraphics[width=\textwidth]{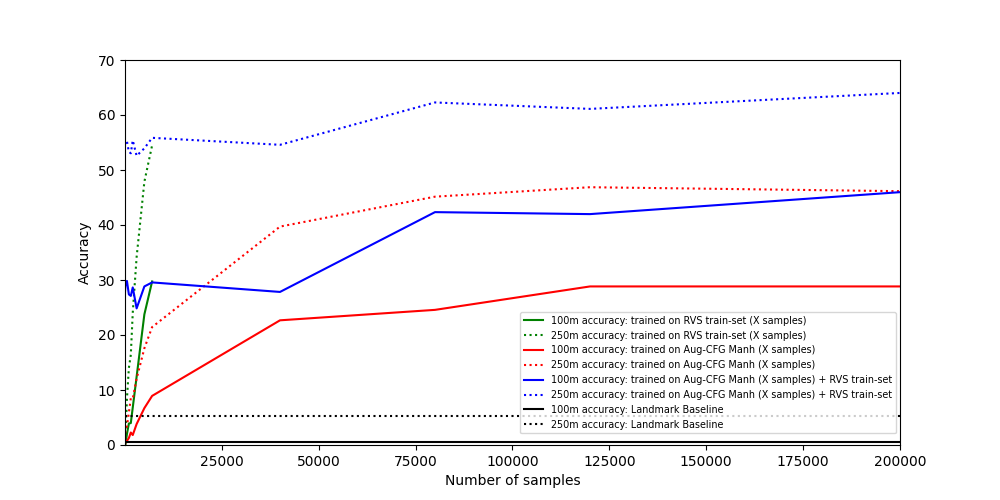}
            \caption[]%
            {{\small Manhattan-dev: Accuracy}}    
            \label{fig:per_number:a}
        \end{subfigure}
        \hfill   
        \begin{subfigure}[b]{0.495\textwidth}  
            \centering 
            \includegraphics[width=\textwidth]{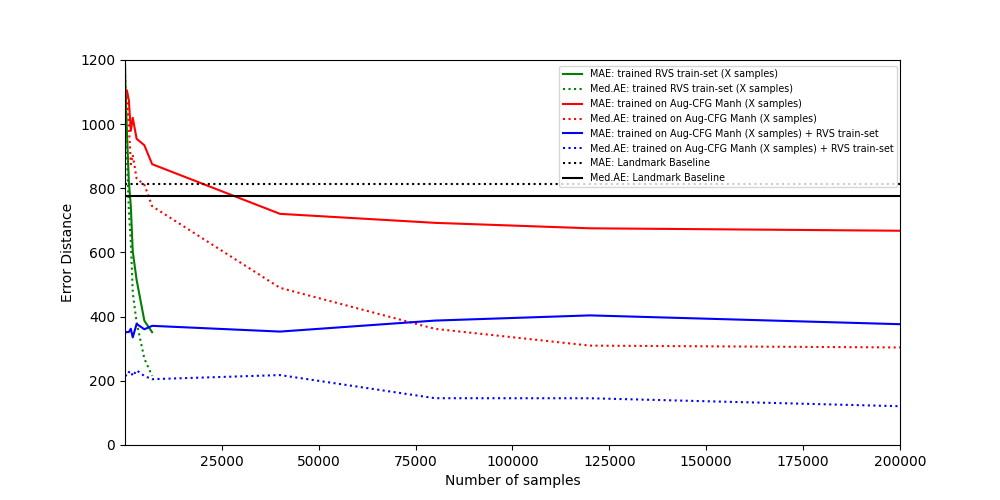}
            \caption[]%
            {{\small Manhattan-dev: Distance Error}}    
            \label{fig:per_number:b}

        \end{subfigure}
        \begin{subfigure}[b]{0.495\textwidth}   
            \centering 
            \includegraphics[width=\textwidth]{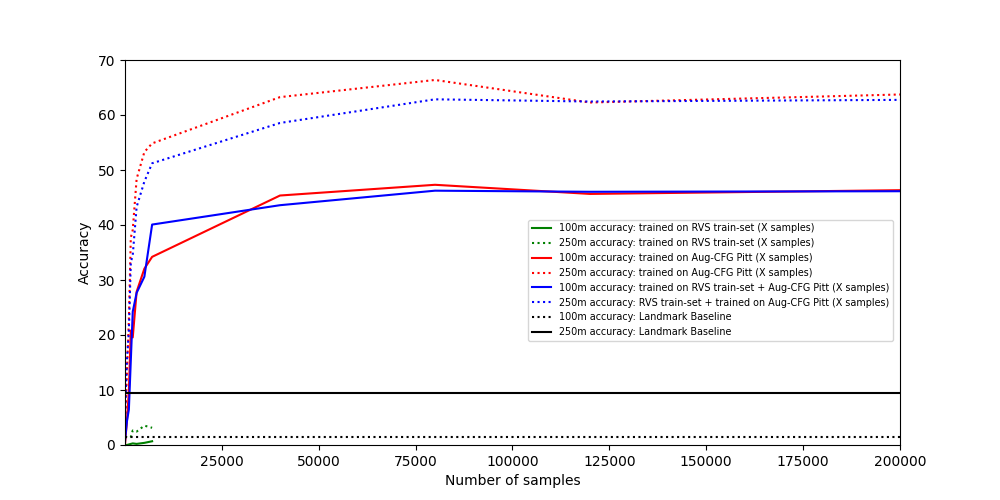}
            \caption[]%
            {{\small Pittsburgh-dev: Accuracy}}    
            \label{fig:per_number:c}
        \end{subfigure}
        \hfill
        \begin{subfigure}[b]{0.495\textwidth}   
            \centering 
            \includegraphics[width=\textwidth]{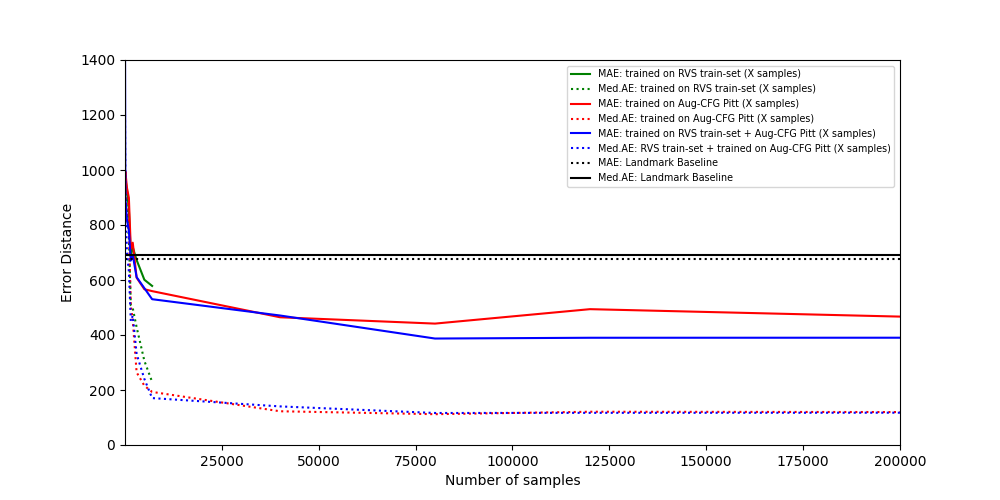}
            \caption[]%
            {{\small Pittsburgh-dev: Distance Error}}    
            \label{fig:per_number:d}
        \end{subfigure}
        \caption
        {T5 performance (Y-axis) with varying AUG-CFG training samples (X-axis). (a,b) RVS seen-city (Manhattan), (c,d) RVS unseen-city (Pittsburgh).} 
        \label{fig:per_number}
    \end{figure*}

Table \ref{tab:results_ablation_CFG} examines variations of CFG-based augmentation. 
Consistent with the emphasis on allocentric spatial relations in RVS, training on Aug-CFG-Allocentric (lines 2, 7) surpasses Aug-CFG-Egocentric (lines 3, 8) in both development sets. 
However, results are mixed regarding whether training on Aug-CFG-Allocentric is better than training on  Aug-CFG. In the seen environment, Aug-CFG-Allocentric (line 2) outperforms Aug-CFG (line 1)  in accuracy but underperforms in error distance. In the unseen-environment (Pittsburgh), the opposite is true --  Aug-CFG-Allocentric (line 7) underperforms Aug-CFG (line 6) in accuracy and overperforms error distance. This inconclusive evidence suggests potential value in investigating a single data approach for tasks with varying demands, such as RUN's reliance on egocentric relations.



Despite capturing identical spatial relations, Aug-CFG-Minimal's limited stylistic variations and vocabulary size due to fewer templates hinder its performance compared to Aug-CFG across both seen (line 1 vs. line 4) and unseen (line 6 vs. line 9) sets. This suggests that the mere presence of accurate spatial relations may not be sufficient for optimal model learning, potentially due to insufficient exposure to diverse linguistic contexts and syntactic structures.


Training on Aug-Dummy teaches the model only the optional paths from a starting point, as the instruction is a dummy one. The poor results achieved over Aug-Dummy in both seen and unseen environments (lines 5, 10), prove that training on Aug-CFG allows the model to learn spatial relations from the instructions. The Aug-Dummy for Pittsburgh results are better as Manhattan is much denser in entities than Pittsburgh, thus, it contains more optional paths from the starting point.

\subsection{Data Quantity Impact}

Figure \ref{fig:per_number} shows four graphs of T5 performance trained on different amounts of AUG-CFG data. The results reveal a quality-quantity trade-off influencing T5's performance.
In seen cities (Figures \ref{fig:per_number:a} and \ref{fig:per_number:b}), 7,000 high-quality human annotations (green lines) outperform 7,000 synthetic AUG-CFG samples (red lines) in 100m accuracy (27.92\% vs. 8.93\%) and Med.AE (231m vs. 744m), indicating that the human-annotated RVS data possesses substantially higher quality than the synthetic data. The steep curve of the RVS train-set (green lines) compared to the mild curve of AUG-CFG (red lines)  further reinforces this conclusion. However, this trend flips with 200K synthetic samples, showcasing the power of quantity over quality when data is abundant. This suggests ample data helps the model grasp the environment and spatial relations.
Additionally, while both RVS train-set (green line) and AUG-CFG (red line) demonstrate strong performance, the combined AUG-CFG + RVS train-set (blue line) exhibits a sustained upward trend, consistently surpassing the green and red lines in both accuracy and distance error. This further indicates that augmenting with a large amount of data can potentially enhance performance beyond even high-quality human annotations. This approach offers a promising solution for the data scarcity challenges often encountered in NLP geospatial tasks.


In the unseen-city split (Figures \ref{fig:per_number:c} and \ref{fig:per_number:d}), the RVS train-set (green line) exhibits steady improvements in accuracy and error distance despite originating from a different region. This demonstrates that high-quality data, even when geographically distinct, can benefit the model.
However, the significant performance gap between the RVS train-set (green) and AUG-CFG (red) lines, even with equal data quantities, reaffirms the importance of regionally-specific data. While the RVS train-set (green) slope nears AUG-CFG (red line), its accuracy improvement remains significantly lagged (moderate slope), suggesting the model primarily learns general directional understanding rather than fine-grained spatial reasoning.
Furthermore, the combined AUG-CFG + RVS train-set (blue) offers only marginal gains over AUG-CFG alone (red), indicating that the benefits of high-quality non-regional data diminish when paired with large-scale augmentation.

\subsection{Distribution Analysis}
\begin{figure}[t]
\scalebox{0.56}{
\hspace{2.9cm}
\includegraphics[width=\textwidth,center]{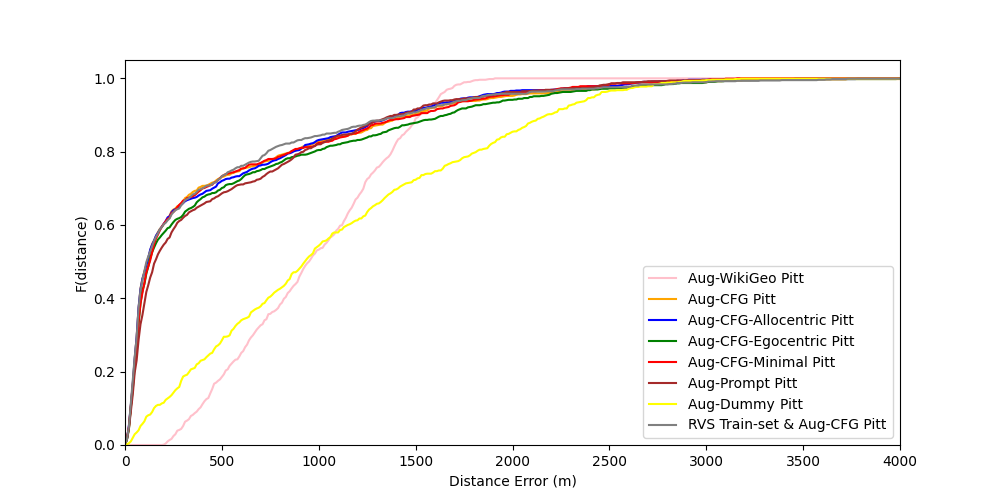}}
 \caption{Cumulative distribution function (CDF) error. 
 Augmentation impact on distance error (meters).
 } 
 \label{fig:distribution}
\end{figure}

Figure \ref{fig:distribution} reports the performance of the various augmentation methods through cumulative distribution functions (CDFs).
These CDFs show the percentage of inferences with error distances below a specific meter value (x-axis), effectively capturing accuracy across all error values and exposing underlying error distributions.
Notably, CFG-based methods (AUG-CFG and its variants) exhibit a remarkably similar distribution, characterized by a sharp accuracy ascent up to approximately 200m. Aug-Dummy's curve (yellow)
resembles a ray emanating from the origin, suggesting a strong correlation with the path distribution assimilated during training. Aug-WikiGeo's low-resolution nature is evident in its CDF (pink). While it starts slow (\~200m) and climbs faster than the yellow line, its limited accuracy holds it back. It only catches up to the CFG-based methods at \~1500m, but then fizzles out (plateaus) before they 
even reach their peak.

\section{Background and Related Work}



Text-based navigation tasks constitute a multimodal challenge \citep{antol2015vqa, zest, ji2022abstract} demanding the integration of language comprehension and environmental knowledge. This environment can be either indoor, commonly represented through the agent's visual perception as it navigates \cite{anderson2018vision}, or outdoor. For outdoor environments, representation falls into three main categories: (i) \emph{Visual Representation} --- similar to indoor settings, real-world imagery can be employed, with agents learning the environment through street-view like exploration \cite{anderson2018vision}; (ii) \emph{Map-based Representation} --- agents navigate based on map perception, as seen in studies utilizing maps as the primary input ~\cite{anderson1991hcrc, RVS}; and (iii) \emph{Hybrid Representation} --- a combination of visual and map information ~\cite{vasudevan2020talk2nav,de2018talk}.

Realistically, we would like to learn to navigate based on text in new environments, ones that our models did not train on before. Furthermore, environments constantly change, such that we need to reacquaint our models with these changes in order for them to stay relevant ~\cite{zhang2021situatedqa}. 

Synthetic data generation, facilitated by LLMs, has become a prominent approach in NLP to mitigate data  \cite{sahu-etal-2022-data, stylianou2023domain}. This technique is particularly attractive due to its adaptability to various tasks and its ability to generate substantial volumes of data, allowing for robust model training and improved performance.
Data generation has also been applied to multimodal tasks, demonstrating performance close to that of human-annotated data ~\cite{bitton2021automatic}.

Previous works on vision and language navigation tasks (VLN) tried to tackle the data scarcity issue by generating synthetic data with a generative model trained with human annotated data ~\cite{fried2018speaker,zhu2020multimodal,majumdar2020improving}. However, the learned distribution was limited to the environment the model was trained on. {\scshape EnvEdit} ~\cite{li2022envedit} tackled the VLN task by generating new environments and synthetic navigation instructions in order to teach models to generalize to new environments. The new environments created differ in style, appearance, or the configuration of the objects. However, a lack of new objects hinders unseen object handling.
\citet{kamath2023new} leveraged image-to-image synthesis  via a Generative Adversarial Network (GAN) architecture  \cite{koh2023simple} to create new viewpoints for existing environments and subsequently generate synthetic instructions for these environments.


Several prior works have tackled unseen environments without employing data generation:
(i) \emph{Entity abstraction} -- \citet{paz2019run} propose learning spatial language independent of specific entities by linking abstracted entities to the agent's perception. However, this approach is limited to simple navigation tasks where the agent has limited perception (i.e., line of sight). It struggles with complex tasks like RVS, which require allocentric spatial reasoning.
(ii) \emph{cross-lingual augmentation} -- \citet{li2022clear} leverage spatial data across different languages for augmentation. However, it still requires some environment-specific annotations, even if not in the target language.
Additionally, geolocation tasks utilizing open-source datasets like Wikipedia often rely on grid-based representations, suffering from spatially unoriented descriptions. This leads to significant retrieval errors exceeding tens of kilometers, limiting their accuracy ~\cite{wing2011simple}. 
(iii) \emph{Graph-based representation} -- representing the environment via a graph and learning the connection between environment and language ~\cite{hegel,RVS}. This approach lacks promising results due to challenges in encoding complex spatial relationships within a graph format.
\citet{schumann} trained a neural network to generate synthetic navigation instructions based on OpenStreetMap representations. However, these instructions were limited to step-by-step, local line-of-sight guidance. Furthermore, the model's performance was evaluated on unseen paths within the same city it was trained on.

This work tackles the problem of generating synthetic instructions for unseen environments, exemplified by the RVS dataset \cite{RVS}. We propose two techniques: an LLM-based approach for its adaptability, and a CFG rule-based approach for increased precision, highlighting the trade-off between data-driven efficiency and human-crafted accuracy.

\section{Conclusion}

This work presents a novel data-augmentation solution for spatial NLP tasks. Leveraging spatial relation extraction 
it generates a vast, albeit slightly less refined, dataset compared to human annotations. 
This quantitative advantage unlocks superior performance, as evidenced by a 44.54\% absolute improvement in 100m accuracy and a 1,170m reduction in median absolute error distance on unseen environments in the RVS dataset.
These results demonstrate the effectiveness of our approach in handling novel environments with sparse or no human data. Moreover, our CFG-based augmentation method offers adherence to correct spatial relations, control over the content, and interpretability, over LLMs in spatial descriptions and could serve as a future tool for evaluation, detection, and mitigation of artifacts such as `hallucinations'.

\section*{Limitations}
\paragraph{Data Dependence} This paper presents a promising data augmentation method for enhancing NLP tasks like outdoor geolocation and navigation. However, its effectiveness hinges on accessing comprehensive geospatial data, which can be difficult to find in indoor ~\cite{anderson2018vision, jain2019stay, nguyen2019vision, thomason2020vision, qi2020reverie, ku2020room} and virtual environments ~\cite{macmahon2006walk,yan2018chalet, misra2018mapping, shridhar2020alfred, kim2020arramon}. Furthermore, the reliance on open-source data introduces the risk of incompleteness or regional unavailability.

\paragraph{Rule-based Scalability vs. LLM-Generated Artifacts}
In addition, the leading augmentation method (CFG-based) used in this study is rule-based, which, while offering control, preciseness, and interpretability, also requires careful rule design. Crafting effective rules can be time-consuming, laborious, and require substantial expertise in the specific domain. This complexity can limit the scalability and adaptability of the method to new situations or contexts. Furthermore, encoding all necessary knowledge into explicit rules can be challenging, especially for complex domains. This can limit the method's ability to capture subtle nuances or unforeseen situations.
This approach also stands in contrast to the prevailing augmentation methods, which are grounded in LLMs \citep{schick2021generating, wang2022less}.
Using LLMs to generate data for augmentation represents a promising avenue but requires further research to address issues like hallucinations, which are particularly critical in spatial descriptions. Therefore, rule-based methods such as our CFG-based method, which are not vulnerable to hallucination problems, could serve as a valuable tool for future research exploring the evaluation, detection, and mitigation of such artifacts in LLM-generated text. Furthermore, by comparing augmentations generated using both methods on specific tasks, we might gain insights into the types of artifacts introduced by LLMs and how rule-based methods can help mitigate them.


\paragraph{Limited Perception: Lack of Visual Cues}
Our study primarily relied on map-based knowledge for navigation tasks, which deviates from how humans navigate in natural settings. Real-world navigation often involves integrating both visual cues and spatial knowledge acquired from maps, a complexity not fully captured in our current approach. 
While the StreetLearn dataset \cite{mirowski2019streetlearn} offers Google Street View imagery for the test environments over both the Manhattan and Pittsburgh regions in the RVS setup, we did not leverage this visual information. Future research could extend the scope of our study by integrating visual perception and map-based knowledge into the text generation approach for augmentation.

\paragraph{Bridging the Human-AI Performance Gap} In spite of progress made in this research, a substantial gulf still separates current models' performance from human performance in the RVS task. Even with our augmentation methods, current models lag behind human performance by 47.55\% and 42.15\% in 100-meter accuracy for seen and unseen environments, respectively. 
Bridging this gap presents a critical challenge and an exciting opportunity for future research, potentially unlocking novel avenues for pushing the boundaries of this task.




\section*{Acknowledgements}

This research has been funded by the European Research Council (ERC), grant number 677352 and by a grant from the Israeli Science Foundation (ISF) number 670/23, for which we are grateful. The research was further supported by
a KAMIN grant from the Israeli Innovation Authority, and computing resources kindly
funded by a VATAT grant and via the Data Science
Institute from Bar-Ilan University (BIU-DSI). 
We are also grateful for the additional support provided by a Google grant.

\bibliography{references}
\bibliographystyle{acl_natbib}

\appendix

\begin{table*}[t]
\scalebox{0.8}{
\begin{tabular}{ll|ll}
 & \textbf{Exaples of  CFG generated templates}                                                                                                                                                                                                                                                      & \textbf{Example of generated instruction from template}                                                                                                                                                                                                                                   &  \\ \cline{2-3}
 & \begin{tabular}[c]{@{}l@{}}“Meet at the END\_POINT. Go CARDINAL\_DIRECTION \\ from MAIN\_PIVOT for INTERSECTIONS intersections. \\ It will be near a NEAR\_PIVOT. If you reach \\ BEYOND\_PIVOT, you have gone too far.”\end{tabular}                                                             & \begin{tabular}[c]{@{}l@{}}“Meet at the library. Go north-east from Starbucks for two \\ intersections. It will be near a book store. If you reach a \\ Zara cloth shop, you have gone too far.”\end{tabular}                                                                             &  \\ \cline{2-3}
 & \begin{tabular}[c]{@{}l@{}}“Go CARDINAL\_DIRECTION from MAIN\_PIVOT for \\ BLOCKS blocks to arrive at the END\_POINT, right \\ GOAL\_POSITION. You will pass MAIN\_NEAR\_PIVOT \\ before reaching the destination. You've overshot the meeting \\ point if you reach BEYOND\_PIVOT.”\end{tabular} & \begin{tabular}[c]{@{}l@{}}“Go east from Grace Church for 3 blocks to arrive at the \\ parking lot, right in the middle of the block. You will pass \\ Jefferson Market Garden before reaching the destination. \\ You've overshot the meeting point if you reach Chipotle.”\end{tabular} &  \\ \cline{2-3}
 & \begin{tabular}[c]{@{}l@{}}“Go CARDINAL\_DIRECTION from MAIN\_PIVOT for \\ BLOCKS blocks to arrive at the END\_POINT, right \\ GOAL\_POSITION. You will see MAIN\_NEAR\_PIVOT \\ before reaching the destination. You've overshot the \\ meeting point if you reach BEYOND\_PIVOT.”\end{tabular}  & \begin{tabular}[c]{@{}l@{}}“Go east from Grace Church for 3 blocks to arrive at the \\ parking lot, right in the middle of the block. You will see \\ Jefferson Market Garden before reaching the destination. \\ You've overshot the meeting point if you reach Chipotle.”\end{tabular}  &  \\ \cline{2-3}
 & \begin{tabular}[c]{@{}l@{}}“Walk CARDINAL\_DIRECTION and past MAIN\_PIVOT \\ to reach the END\_POINT. The END\_POINT is not far \\ from NEAR\_PIVOT.”\end{tabular}                                                                                                                                & \begin{tabular}[c]{@{}l@{}}“Walk north and past Washington Square Park to reach \\ the cafe. The cafe is not far from a tobacco shop.”\end{tabular}                                                                                                                                       &  \\ \cline{2-3}
 & \begin{tabular}[c]{@{}l@{}}“Head to MAIN\_PIVOT and go CARDINAL\_DIRECTION \\ and meet at the END\_POINT, right GOAL\_POSITION.”\end{tabular}                                                                                                                                                     & \begin{tabular}[c]{@{}l@{}}“Head to Washington Square Park and go north and meet \\ at the cafe, right on the southeast corner of the block.”\end{tabular}                                                                                                                                &  \\ \cline{2-3}
                                                                                                                                                                                                                                                                                    
\end{tabular}
}

\caption{Examples of navigation instructions and the Context-Free Grammar (CFG)-derived templates they are created from. }
\label{tab:multiple_cfg_examples}
\end{table*}

 \begin{figure*}[h]
   \centering
\scalebox{0.91}{
            \includegraphics[width=1 \textwidth]{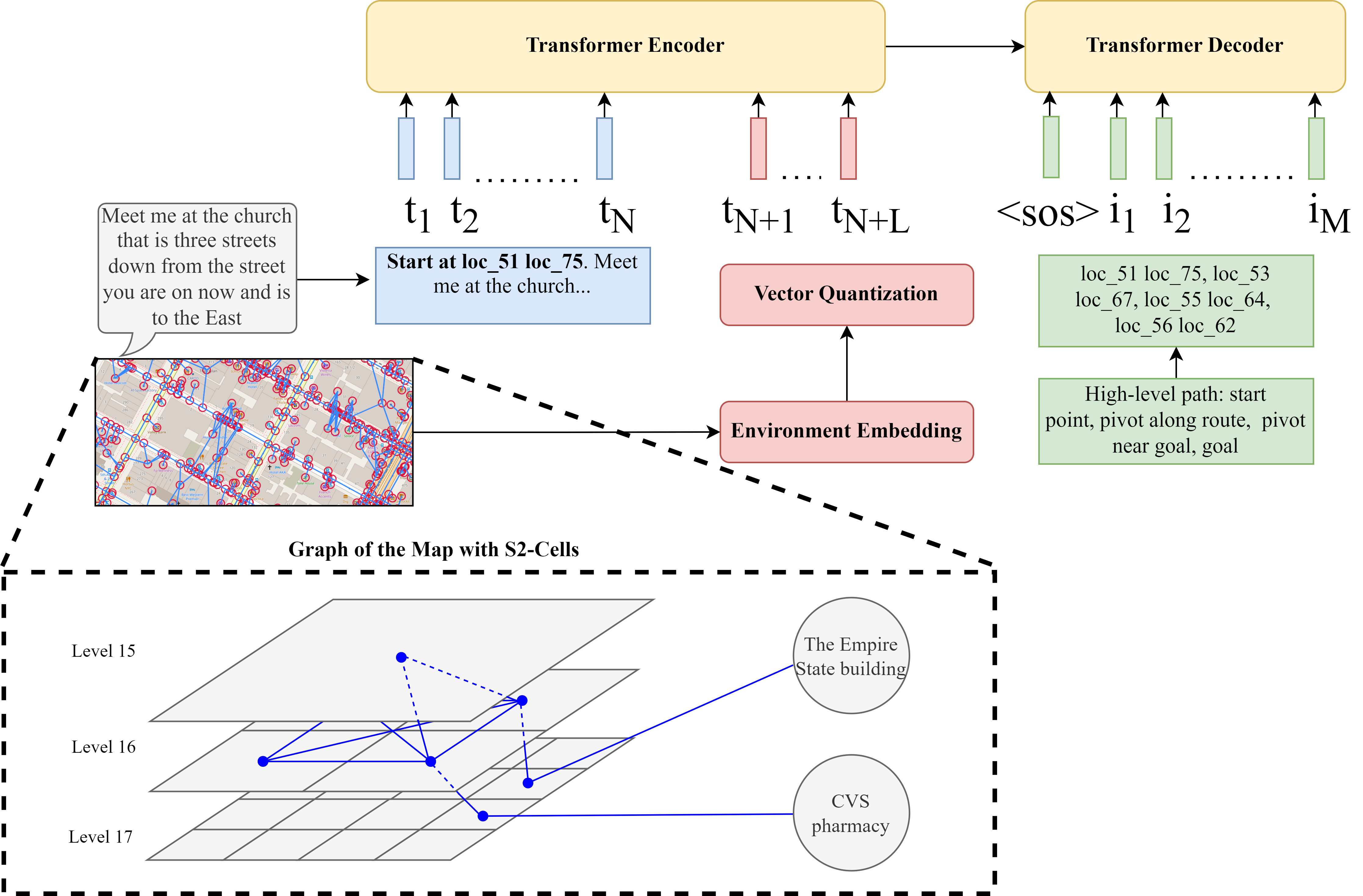}}
            
            \vspace*{-1mm}
                 {\scriptsize{ \begin{spacing}{0.5}

\end{spacing} }}

        \caption
        {The RVS model based on a T5 transformer and a graph representation of the environment \cite{RVS}. } 

        \label{fig:model}
    \end{figure*}

\section{Data Generation}
Our CFG rules define navigation instructions by combining five key elements in various orders:
\begin{itemize}
  \item \textbf{Goal Description:} This specifies the target location.
  \item \textbf{Main Path:} It outlines the primary route using landmarks.
  \item \textbf{Approaching Goal:} This details how to get close to the target.
  \item \textbf{Goal Landmarks:} It describes landmarks relative to the final position.
  \item \textbf{Off-Path Awareness:} This identifies elements to avoid while navigating.
\end{itemize}

Table \ref{tab:multiple_cfg_examples} demonstrates five navigation instructions and their corresponding CFG-templates.

\section{Models}
\label{app_T5_graph}

We also test on the {\scshape T5+Graph} presented in RVS, which is a T5 model incorporating an OSM-based graph representation of the environment. 


\subsection{The Graph Representation}
 A location can be represented by its position ({\em where} the location is) or by its semantics ({\em what} is present at the location, e.g., `a bar'). Semantic knowledge is crucial for grounding mentioned entities to their physical references in the environment. To this end, we aim to represent the semantics via the RVS map-graph. 
We use the RVS map-graph and connect each node to its corresponding S2-cells. As the S2-geometry is a hierarchical structure, we allow for multiple levels of S2-cells connections. Also there are edges between neighboring S2-cells at a given level (see bottom part in Figure \ref{fig:model}). 
To learn an embedding for each S2-cell in the environment, we compute random walks on the graph using node2vec algorithm \citep{grover2016node2vec}.
Following \citet{yu2021vector}, we use linear projection to cluster the graph embeddings into K categories  using the k-means algorithm with cosine similarity distance.  A new token is assigned to each  category and added to the tokenizer's vocabulary. We perform multiple clusters and pass the graph's tokens with the instruction's tokens to the transformer encoder.

\subsection{Experimental Setup Details}

\paragraph{The Graph Embedding} The graph was constructed using three levels of S2-Cells: 15, 16, and 17. At level 16, each sub-graph consisting of four neighboring S2-Cells was fully connected. All S2-Cells in the graph were linked to their parent S2-Cell based on the S2-geometry's hierarchy (i.e., level 17 S2-Cells were connected to level 16 S2-Cells and level 16 S2-Cells were connected to level 15 S2-Cells). Extracted entities from OSM and  \href{https://www.wikidata.org}{Wikidata}
were linked to the smallest level 17 S2-Cell that encompassed their geometry. The node of the entity included additional data such as their geometry, type and name of entity. Random walks on the graph were performed using node2vec \citep{grover2016node2vec}.

\begin{table*}[t]
\scalebox{0.61}{

\begin{tabular}{llccccccc}
& \textbf{Method}   & \textbf{Training Set}   &  \textbf{100m Accuracy} & \textbf{250m Accuracy}         & \textbf{MAE}                                        & \textbf{Med.AE} & \textbf{Max.AE}  & \textbf{AUC} \\ 

\cline{2-9} 
&
\multicolumn{8}{c}{\cellcolor[HTML]{EFEFEF}\textbf{Manhattan (Manh) Seen-city Development Results}}                                                                                                                         \\ \cline{2-9}

{\tiny 1} & \textbf{{\scshape T5} }  & RVS Train-set      & \pzz\pzz 27.92 \small{(0.39)}           
& \pzz\pzz52.63 \small{(0.45)}
& \pzz\pzz\pz 362 \small{(9)}                                             & \pzz\pzz \pz231 \small{(3)}       & \pzz\pzz2,957 \small{(641)} & \pzz\pzz0.32 \small{(0.00)}           \\

{\tiny 2} & \textbf{{\scshape T5+Graph}} & RVS Train-set  & \pzz\pzz29.40 \small{(1.18)}         
& \pzz\pzz54.67  \small{(1.04)}                  

& \pzz\pzz\pz \textbf{357} \small{(7)}                                             & \pzz\pzz \pz216 \small{(8)}       & \pzz\pzz3,889 \small{(826)}   & \pzz\pzz0.31 \small{(0.01)}           
\\ \hdashline

{\tiny 3} & \textbf{{\scshape T5}} & Aug-CFG Manh & \pzz\pzz 28.83 \small{(0.63)}         
& \pzz\pzz 46.15  \small{(0.77)}                  

& \pzz\pzz\pz 668 \small{(17)}                                             & \pzz\pzz \pz 304 \small{(27)}       & \pzz\pzz \pz 4,637 \small{(2,207)}   & \pzz\pzz 0.34 \small{(0.00)} 
\\

{\tiny 4} & \textbf{{\scshape T5+Graph}} & Aug-CFG Manh & \pzz\pzz 30.25 \small{(0.95)}         
& \pzz\pzz  48.11 \small{(0.92)}                

& \pzz\pzz\pz 660 \small{(24)}                                          & \pzz\pzz\pz 299 \small{(17)}       & \pzz\pzz 4,447 \small{(677)}   & \pzz\pzz 0.34 \small{(0.01)}           \\

{\tiny 5} & \textbf{{\scshape T5}} & Aug-CFG Manh \& RVS Train-set & \pzz\pzz \textbf{45.97 \small{(1.34)}    }     
& \pzz\pzz \textbf{64.01  \small{(0.89)}  }                

& \pzz\pzz\pz 377 \small{(32)}                                            & \pzz\pzz \pz \textbf{121 \small{(15)} }      & \pzz\pzz  5,317 \small{(831)}   & \pzz\pz 0.3 \small{(0.00)}

\\

{\tiny 6} & \textbf{{\scshape T5+Graph}} & Aug-CFG Manh \& RVS Train-set & \pzz\pz 45.42 \small{(0.9)}         
& \pzz 63.1  \small{(1.41)}                

& \pzz\pzz\pz 388 \small{(10)}                                          & \pzz\pzz 131 \small{12)}       & \pzz\pz 3,162 \small{(43)}   & \pzz\pz 0.3 \small{(0.01)}           \\

\cline{2-9} 
&
\multicolumn{8}{c}{\cellcolor[HTML]{EFEFEF}\textbf{Pittsburgh (Pitt) Unseen-Development Results}}                                                                                                                      \\ \cline{2-9} 

{\tiny 7} & \textbf{T5}    & RVS Train-set    & \pzz\pzz\pz 0.49 \small{(1.47)}            &               \pzz\pzz  2.34 \small{(1.44)}                 & \pzz\pzz1,171 \small{(24)}                                                 & \pzz \pz1,107 \small{(14)}            & \pzz\pzz4,701 \small{(101)}         & \pzz\pzz 0.41 \small{(0.00)}           \\

{\tiny 8} & \textbf{{\scshape T5+Graph}} & RVS Train-set  & \pzz\pzz\pz 0.49 \small{(1.01)}            &               \pzz\pzz  2.91 \small{(1.37)}                    & \pzz\pzz 1,067 \small{(77)}                                                & \pzz \pz1,039 \small{(56)}            & \pzz\pzz4,102 \small{(727)}         & \pzz\pzz0.40 \small{(0.00)}           \\ 
 \hdashline

{\tiny 9} & \textbf{{\scshape T5}} & Aug-CFG Pitt & \pzz\pzz {46.63} \small{(0.54)}         
& \pzz\pz {63.73}  \small{(0.41)}                  

& \pzz\pzz 466 \small{(5)}                                             & \pzz\pz 120 \small{(1)}       & \pzz5,251 \small{(0)}   & \pzz\pzz0.31 \small{(0.00)}           \\

{\tiny 10} & \textbf{{\scshape T5+Graph}} & Aug-CFG Pitt & \pzz\pzz \textbf{46.67} \small{(1.45)}         
& \pzz 63.1  \small{(1.59)}                

& \pzz\pzz 474 \small{(1)}                                          & \pzz\pz 119 \small{(8)}       & \pzz 5,251 \small{(0)}   & \pzz\pzz0.31 \small{(0.00)}           \\

{\tiny 11} & \textbf{{\scshape T5}} & RVS Train-set \& Aug-CFG Pitt & \pzz\pzz 46.24 \small{(0.30)}         
& \pzz\pz 62.85  \small{(0.41)}                

& \pzz\pzz\pz \textbf{387} \small{(17)}                                             & \pzz\pz \textbf{116} \small{(4)}       & \pzz\pzz 5,162 \small{(103)}   & \pzz\pzz0.30 \small{(0.00)}           \\

{\tiny 12} & \textbf{{\scshape T5+Graph}} & RVS Train-set \& Aug-CFG Pitt & \pzz\pzz 45.75 \small{(0.62)}         
& \pz \textbf{63.93}  \small{(0)}                

& \pzz\pzz 467 \small{(7)}                                          & \pzz\pz 125 \small{(1)}       & \pzz\pzz 6,509 \small{(755)}   & \pzz\pzz 0.31 \small{(0.00)}          
\end{tabular}
}
\caption{T5+Graph Results: 
Results are divided over RVS's development sets. The augmentations data used for training depends on the method and region that corresponds to the evaluation region: Manhattan (Manh) and Pittsburgh (Pitt).
The distance errors are presented in meters. For the learning models, we report the mean over three random initializations and the standard deviation (STD) is in brackets.  }
\label{tab:results_ablation_T5_graph}

\end{table*}

For both T5-base models we use a pre-trained `T5-Base' model from \href{https://huggingface.co/transformers/v3.0.2/_modules/transformers/modeling_tf_t5.html#TFT5ForConditionalGeneration}{Hugging Face Hub}, which is licensed under the Apache License 2.0. The T5 model was trained on the Colossal Clean Crawled Corpus (C4, \citet{raffel2020exploring}).
The cross-entropy loss function was optimized with
AdamW optimizer \citep{adamW}. The hyperparameter tuning is based on the average results
run with three different seeds. We used a learning rate
of 1e-4. The S2-cell level was searched in [15,
16, 17, 18] and 16 was chosen. The number of clusters for the quantization process was searched in [50, 100, 150, 200, 250] and 150 was chosen. We used 2 quantization layers. Number of epochs for
early stopping was based on their average learning
curve. 
We used the following parameters for the node2vec algorithm: an embedding size of 1024, a walk length of 20, 200 walks, a context window size of 10, a word batch of 4, and 5 epochs.

\section{Additional Results}

Table \ref{tab:results_ablation_T5_graph} demonstrates the performance of {\scshape T5+Graph} with CFG augmentation. The {\scshape T5+Graph} model’s performance is improved by the CFG augmentation in both seen and unseen environments (lines 2 vs 4, and lines 8 vs 10). In the seen environment, {\scshape T5+Graph} with CFG augmentation achieves higher scores than T5 with CFG augmentation (lines 3 vs 4). However, there is no clear evidence to suggest which model performs better when using CFG augmentation data in both seen and unseen environments. This finding suggests that the explicit spatial knowledge offered by CFG augmentation data provides an easier path to learn spatial relations, making the graph information in {\scshape T5+Graph} redundant or even detrimental.

\end{document}